\documentclass[sigconf, nonacm]{acmart}

\AtBeginDocument{%
  }

\usepackage{balance}
\copyrightyear{2027}
\acmYear{2027}

\setcopyright{none}
\renewcommand\footnotetextcopyrightpermission[1]{}
\begin{document}

\title{AffectiveVLA: Safe and Intent-Aware VLA Control for Human–Humanoid Social Interaction}

\author{Nikita Kuzmin}
\affiliation{%
  \institution{Skolkovo Institute of Science and Technology}
  \city{Moscow}
  \country{Russia}}
\email{Nikita.Kuzmin@skoltech.ru}

\author{Georgii Demianchuk}
\affiliation{%
  \institution{Skolkovo Institute of Science and Technology}
  \city{Moscow}
  \country{Russia}}
\email{Georgii.Demianchuk@skoltech.ru}

\author{Mariya Lezina}
\affiliation{%
  \institution{Skolkovo Institute of Science and Technology}
  \city{Moscow}
  \country{Russia}}
\email{Mariya.Lezina@skoltech.ru}

\author{Yara Mahmoud}
\affiliation{%
  \institution{Skolkovo Institute of Science and Technology}
  \city{Moscow}
  \country{Russia}}  
\email{Yara.Mahmoud@skoltech.ru}

\author{Iana Zhura}
\affiliation{%
  \institution{Skolkovo Institute of Science and Technology}
  \city{Moscow}
  \country{Russia}}
\email{Iana.Zhura@skoltech.ru}

\author{Miguel Altamirano Cabrera}
\affiliation{%
  \institution{Skolkovo Institute of Science and Technology}
  \city{Moscow}
  \country{Russia}}
\email{m.altamirano@skoltech.ru}

\author{Dzmitry Tsetserukou}
\affiliation{%
  \institution{Skolkovo Institute of Science and Technology}
  \city{Moscow}
  \country{Russia}}
\email{d.tsetserukou@skoltech.ru}

\renewcommand{\shortauthors}{Kuzmin et al.}

\begin{abstract}
Social touch interactions, such as handshakes, fist bumps, and hugs, require humanoid robots to infer human intent, generate contact-rich motions, and maintain safe physical contact. We present AffectiveVLA, an intent-gated Vision-Language-Action (VLA) pipeline deployed on a Unitree G1 humanoid with dexterous hands. A lightweight RGB-based model recognizes interaction intent from two onboard cameras and converts it into a language prompt for the VLA policy. Low-confidence predictions suppress motion execution, reducing unintended activations. During physical contact, interaction-specific compliance profiles adjust arm impedance, while an independent voice-triggered mechanism initiates a latched safe-stop procedure. Using VR teleoperation, we collected approximately 1,300 multimodal interaction episodes and fine-tuned four VLA models: GR00T, RLDX-1, Being-H0.5, and UnifoLM-VLA-0. On-robot evaluation identifies GR00T as the best-performing standalone model. Integrating intent recognition increases its macro-F1 score from 0.75 to 0.90, while the intent model achieves a p95 inference latency of 6.27 ms. These results demonstrate that explicit intent gating improves the reliability of VLA-based social interaction without compromising real-time operation.
\end{abstract}


\begin{teaserfigure}
 \includegraphics[width=\textwidth]{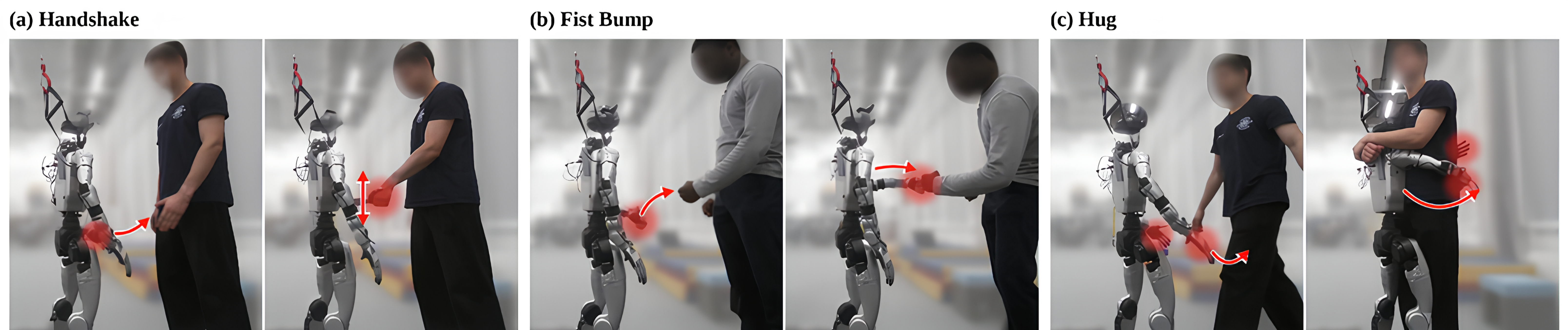}
   \caption{AffectiveVLA enables intent-aware and contact-safe execution
  of three social interactions: (a) handshake, (b) fist bump, and
  (c) hug. Arrows indicate the direction of interaction motion, and
  red regions highlight physical contact.}
  \Description{A sequence of photographs showing a humanoid robot
  interacting with people through a handshake, a fist bump, and a hug.
  Red highlights mark contact regions and arrows indicate motion.}
  \label{fig:teaser}
\end{teaserfigure}


\maketitle
\section{Introduction} 
As humanoid robots enter human environments, they must participate in social rituals like handshakes, hugs, and fist bumps (Fig.~\ref{fig:teaser}). Unlike object manipulation, these interactions involve an active human partner whose intent must be inferred, requiring the robot to meet a moving target while ensuring safe, comfortable contact~\cite{block2021six,block2023arms}. 

Existing approaches address this only partially: learned whole-body policies rely on explicit operator commands~\cite{huang2026beyond}, compliance controllers typically use fixed settings~\cite{lu2025gentlehumanoid}, and vision-language pipelines infer intent with high latency, lacking physical execution~\cite{bao2025hierarchical}. While end-to-end Vision-Language-Action (VLA) models can generate contact-rich motions, they risk unsafe false activations when human intent is uncertain.

To address this, we propose a four-component pipeline: (1) RGB-based intent recognition to gate and prompt the VLA, (2) VLA-based whole-body motion generation, (3) task-specific compliance adaptation, and (4) an independent voice-triggered safe stop.

This work investigates the following research questions:
\begin{itemize}
    \item \textbf{RQ1:} Which VLA model best performs contact-rich HRI actions
    after fine-tuning on the same dataset?
    \item \textbf{RQ2:} Does the preferred arm compliance depend on the type
    of social interaction?
    \item \textbf{RQ3:} Does intent gating improve the selection of the correct interaction, and can ongoing contact be safely interrupted?
\end{itemize}

Our main contributions are:
\begin{itemize}
    \item A comparative evaluation of four fine-tuned VLA models (Being-H0.5, GR00T, RLDX-1, UnifoLM-VLA-0) on social touch tasks .
    \item An intent-gated VLA pipeline with task-conditioned compliance and voice-triggered safe stopping, validated via ablation and user studies.
\end{itemize}

\section{Related Work}

\paragraph{Social touch and learning-based interaction.}
Social touch in HRI has focused on single, pre-selected interactions: HuggieBot established design guidelines for hugging robots~\cite{block2021six,block2023arms}, and tactile sensing has been used to time handshake phases~\cite{pequeno2024tactile}. Reinforcement learning now scales to general-purpose whole-body controllers~\cite{luo2025sonic}, with systems like ECHO~\cite{mascaro2024robot}, RHINO~\cite{chen2025rhino}, and D-STAR~\cite{huang2026beyond} learning hug and handshake policies. However, these systems rely on fixed skill sets and operator-selected interaction types rather than inferring intent from the partner. For safe contact, impedance control allows the robot to yield~\cite{lu2025gentlehumanoid}, but most systems use fixed compliance settings. For intent inference, foundation models can reason about social cues~\cite{kim2024understanding}, but VLM-based approaches like HIAER~\cite{bao2025hierarchical} suffer from multi-second latency and cannot execute contact-rich gestures.

\paragraph{Vision-language-action models.}
VLA models map images and language to robot actions~\cite{gr00tn1_2025}, extending to humanoid control~\cite{xue2025leverb}, cross-embodiment learning~\cite{beingbeyond2026beingh05}, and dexterous manipulation~\cite{rldx2026,unifolm-vla-0}. These models target manipulation rather than social contact, though GIVE~\cite{liu2026give} shows that textual gesture descriptions improve intent grounding. We use GR00T~N1.6~\cite{gr00tn1_2025}, RLDX-1~\cite{rldx2026}, Being-H0.5~\cite{beingbeyond2026beingh05}, and UnifoLM-VLA-0~\cite{unifolm-vla-0} as interchangeable action backbones.

\paragraph{Positioning.}
Unlike skill-specific policies~\cite{huang2026beyond}, we infer interaction from dual RGB views and verbalize it as a VLA prompt, enabling new gestures through data rather than reward design. Unlike HIAER~\cite{bao2025hierarchical}, intent is recognized in $\sim$6\,ms by a lightweight video model, and the robot performs physical contact. Unlike fixed-compliance controllers~\cite{lu2025gentlehumanoid}, impedance adapts to each interaction. To our knowledge, this is the first system combining RGB-based intent recognition with VLA execution and interaction-specific impedance for social touch on a humanoid.

\section{System and Dataset}

\subsection{Robot and Teleoperation Setup}
\label{sec:setup}

The platform used is a Unitree G1 humanoid with BrainCo dexterous hands, controlled by the SONIC whole-body controller, which retargets operator motion into joint commands with per-joint impe-dance gains ($k_p$, $k_d$) for 17 upper-body motors. Teleoperation uses a PICO~4 Ultra headset running XRoboToolkit~\cite{zhao2026xrobotoolkit}, streaming head/controller poses and 26-joint hand tracking at 90\,Hz with stereoscopic passthrough feedback. Upper-body commands are forwarded to SONIC and finger trajectories to the BrainCo hands at 50\,Hz, with grip presses serving as the episode-control interface. Each episode records three synchronized RGB views --- ego-centric ($1280{\times}720$) and head-mounted ($1600{\times}896$) on the robot at 15\,FPS, plus an external scene view ($1280{\times}960$) on the host --- and upper-body telemetry (joint states, velocities, torques, commands) at 50\,Hz.

\subsection{Dataset Collection and Cleaning}
\label{sec:dataset}

Demonstrations were collected by teleoperating the robot with a human partner in four interaction classes: \emph{handshake}, \emph{fist bump}, \emph{hug}, and \emph{no interaction}. Each episode stores participant profile, multimodal observations, robot states, action commands, and interaction labels in LeRobot format. Retained episodes were annotated with four temporal markers --- \emph{approach start}, \emph{contact start}, \emph{release start}, and \emph{idle start} --- used to train the interaction-intent model (Sec.~\ref{sec:human_intent_recognition}). Episodes with tracking loss, safety stops, unintended contacts, or anomalous duration were discarded. The final dataset contains 1,270 episodes from 19 participants (13 male, 6 female; age $24.4 \pm 3.7$ years; height $175.4 \pm 8.6$ cm), with no participant overlap between the extended set (982 episodes, 15 participants) and test set (288 episodes, 4 unseen participants).

\subsection{VLA Models}
\label{sec:vla}

We evaluate four VLA models: Being-H0.5~\cite{beingbeyond2026beingh05}, GR00T~N1.6~\cite{gr00tn1_2025}, RLDX-1~\cite{rldx2026}, and UnifoLM-VLA-0~\cite{unifolm-vla-0}. All follow the same architecture --- a ViT-based VLM backbone, a projector aligning visual and language features, and an action expert predicting chunked actions --- operating in a unified 76-dimensional action space: 64-dimensional SONIC motion tokens for the upper body plus 6-DoF finger commands for each BrainCo hand, chunked over 40 steps. All models are fine-tuned using model-specific recipes. For Being-H0.5, RLDX-1, and UnifoLM-VLA-0, the language backbone is frozen while the ViT encoder, projector, and action expert are trained. For GR00T~N1.6, only the action expert and projector are updated. Fine-tuning runs on a single A100 80GB GPU for 30k steps; UnifoLM additionally requires converting the dataset to HDF5/RLDS format.

\section{Human-Aware VLA Pipeline}

\begin{figure*}[t]
  \centering
  \includegraphics[width=0.85\textwidth]{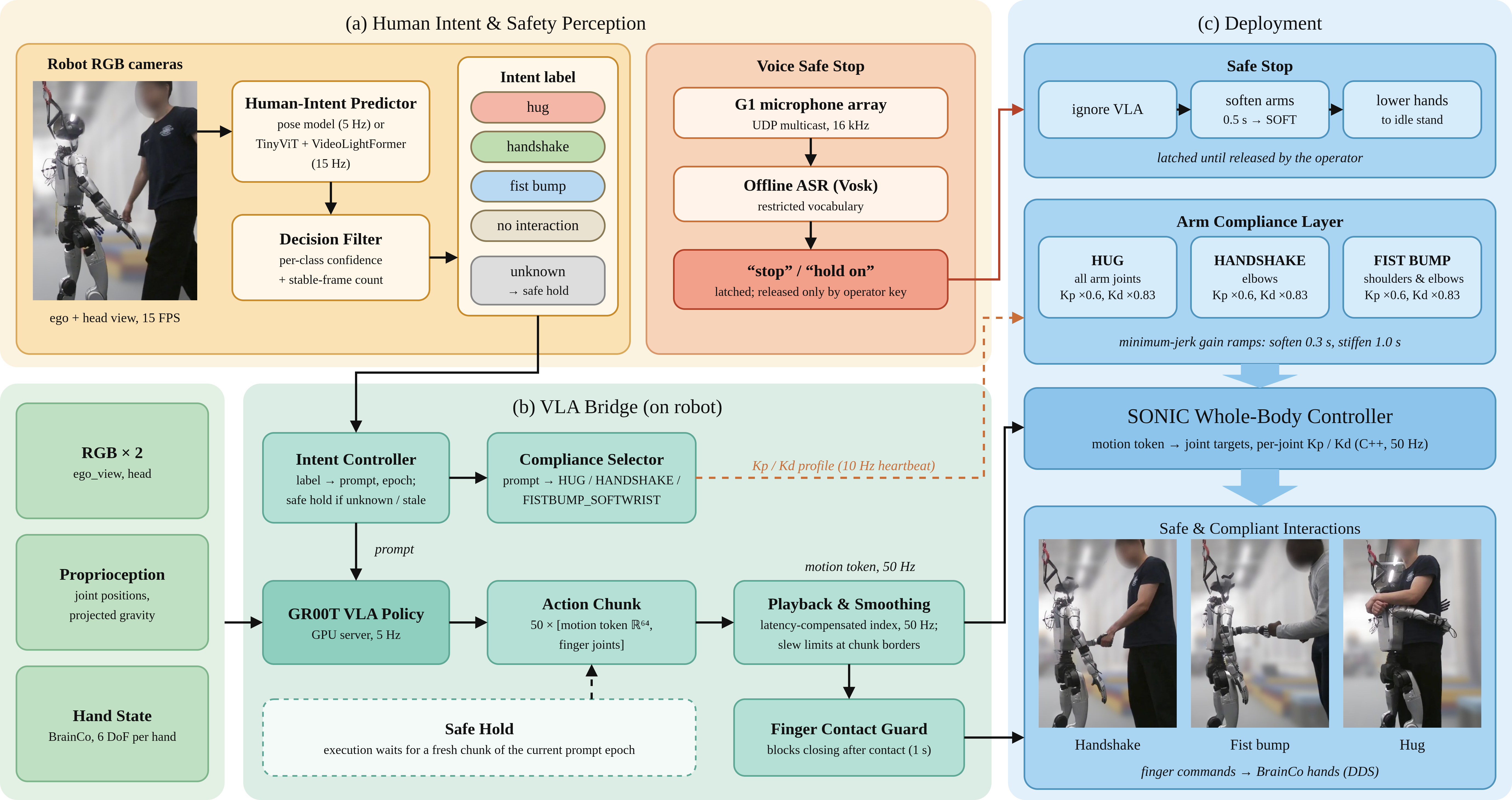}
  \caption{Overview of the proposed human-aware VLA pipeline.
  (a) The perception module estimates the human partner's interaction intent from two RGB camera views, filters uncertain predictions, and independently detects voice commands for safe stopping.
  (b) The on-robot VLA bridge converts a confirmed intent into a task prompt, predicts whole-body action chunks, and applies playback smoothing and finger-contact safeguards.
  (c) During deployment, task-conditioned arm-compliance profiles and the latched safe-stop mechanism supervise the SONIC whole-body controller,
  enabling safe and compliant handshakes, fist bumps, and hugs.}
  \Description{A three-part system diagram. Human intent is estimated from two robot camera views, while a separate voice module detects stop commands.
  Confirmed intent is converted into a prompt for the VLA policy, which
  generates smoothed whole-body and finger actions. During deployment,
  interaction-specific arm compliance and a latched safe stop supervise
  execution of handshakes, fist bumps, and hugs.}
  \label{fig:pipeline}
\end{figure*}

Figure~\ref{fig:pipeline} presents the proposed pipeline. Camera observations
are used to recognize interaction intent. A confirmed intent is converted into
a task prompt for the VLA model, while task-specific compliance and an
independent voice-triggered safe-stop mechanism supervise physical execution.

\subsection{Human Intent Recognition}
\label{sec:human_intent_recognition}

The human-intent module recognizes the interaction initiated by the human from two synchronized onboard RGB views: one for full-body motion and one for the hand-contact area. At each inference step, it processes the 16 most recent frame pairs sampled at 15 Hz, corresponding to about one second of motion history. Frames are resized to $224\times224$ using letterboxing.

A shared TinyViT encoder extracts frame-level features from both views. Separate causal VideoLightFormer modules aggregate temporal information, and attention-based fusion combines the two representations. The model classifies each observation window as \texttt{handshake}, \texttt{hug}, \texttt{fist\_bump}, or \texttt{no\_interaction}.

The recognized interaction is converted into the task prompt for the VLA policy. An intent is committed only when its probability exceeds 0.6 for three consecutive inference steps. Until then, motion execution remains disabled, reducing unintended activations from uncertain predictions.

\subsection{VLA Whole-Body Execution}

Once an intent is accepted, it is mapped to a task prompt such as \textit{hug}, \textit{handshake}, or \textit{fist bump}. The corresponding compliance profile is activated, and the prompt is passed to the VLA together with two camera views, the robot state, and measured BrainCo hand states.

The VLA runs at \(5\,\mathrm{Hz}\) and outputs action chunks with 64-dimensional motion tokens and normalized commands for the six actuators of each BrainCo hand. Actions are executed at \(50\,\mathrm{Hz}\). Motion tokens are passed to GEAR-SONIC for whole-body joint targets and balance control, while finger commands are sent directly to the BrainCo controllers.

Per-actuator rate limiting prevents abrupt changes between chunks. Contact is detected from the difference between commanded and measured finger positions; after contact, further closing is blocked while opening remains enabled. If the prompt changes, queued actions are discarded and execution is held until a matching chunk is received. The compliance profile is maintained by a \(10\,\mathrm{Hz}\) heartbeat; if it stops, the last gains are held.

\subsection{Task-Conditioned Compliance}
The compliance layer scales the impedance gains of the 14 arm joints, leaving torso and leg gains unchanged:
\begin{equation}
K_{p,i} = \alpha_i K^{0}_{p,i}, \qquad K_{d,i} = \beta_i K^{0}_{d,i},
\label{eq:gains}
\end{equation}
with $\beta_i \geq \sqrt{\alpha_i}$. The recognized intent selects the profile; see Table \ref{tab:compliance_profiles}. Profiles change along minimum-jerk ramps (0.3\,s softening, 1.0\,s stiffening) that keep damping high during each transition.

\begin{table}[t]
    \centering
    \caption{Implemented compliance profiles, shown as ($K_p$ scale, $K_d$ scale). Interaction profiles are provisional, chosen from pilot sessions; final values will be set by the user study (Sec.~\ref{sec:exp_compliance}).}
    \label{tab:compliance_profiles}
    \begin{tabular*}{\columnwidth}{@{\extracolsep{\fill}}lccc@{}}
        \toprule
        Profile & Shoulder & Elbow & Wrist \\
        \midrule
        \textsc{Handshake}
            & \((1.00,1.00)\)
            & \((0.60,0.83)\)
            & \((1.00,1.00)\) \\
        \textsc{Hug}
            & \((0.60,0.83)\)
            & \((0.60,0.83)\)
            & \((0.60,0.83)\) \\
        \textsc{FistBump}
            & \((0.60,0.83)\)
            & \((0.60,0.83)\)
            &  \((1.00,1.00)\) \\
        \textsc{Soft}
            & \((0.25,0.50)\)
            & \((0.25,0.50)\)
            & \((0.25,0.50)\) \\
        \bottomrule
    \end{tabular*}
\end{table}
\subsection{Voice-Triggered Safe Stop}
An offline Vosk recognizer~\cite{vosk} with a restricted vocabulary (``stop'', ``hold on'') triggers a deterministic, latched stop: operator and VLA commands are ignored, locomotion is set to idle, the arms soften to \textsc{SOFT} and are then lowered to the rest pose (first outward if a hug is detected), and the BrainCo fingers open. Only a manual release resumes control.

\section{Experiments}
\label{sec:experiments}
\subsection{Offline VLA Comparison}
\label{sec:exp_offline}

Each model (30k fine-tuning steps) predicts 40-step action chunks along the 288 held-out test episodes, and the predictions are compared with the teleoperated ground truth (Table~\ref{tab:vla_offline}). Being-H0.5, GR00T, and RLDX-1 are within 2\% of each other in overall RMSE, while UnifoLM-VLA-0 is worse on all metrics. GR00T is the most accurate at the first and last chunk steps. Since the VLA is queried at 5~Hz and executed at 50~Hz, mainly the beginning of each chunk is executed, so first-step accuracy matters most on the robot.

\begin{table}[t]
\centering
\caption{Offline open-loop action error (RMSE, lower is better) on 288 test
episodes with 40-step action chunks. \emph{Body}: 64-D SONIC motion token;
\emph{Hand}: BrainCo finger commands; \emph{First}/\emph{Last}: error at the
first and last step of the chunk. Best in bold.}
\label{tab:vla_offline}
\small
\begin{tabular*}{\columnwidth}{@{\extracolsep{\fill}}lccccc@{}}
\toprule
 & & \multicolumn{2}{c}{Action part} & \multicolumn{2}{c}{Chunk step} \\
\cmidrule(lr){3-4}\cmidrule(l){5-6}
Model & Overall & Body & Hand & First & Last \\
\midrule
Being-H0.5    & 0.135 & \textbf{0.100} & 0.353 & 0.093 & 0.161 \\
GR00T         & 0.135 & 0.103 & \textbf{0.341} & \textbf{0.085} & \textbf{0.156} \\
RLDX-1        & \textbf{0.133} & \textbf{0.100} & \textbf{0.341} & 0.096 & 0.158 \\
UnifoLM-VLA-0 & 0.148 & 0.114 & 0.373 & 0.110 & 0.177 \\
\bottomrule
\end{tabular*}
\end{table}

\subsection{Online Evaluation on the Robot}
\label{sec:exp_online}

Five participants performed each interaction ten times ($5{\times}3{\times}10 = 150$ trials per policy) with each VLA alone and with our system (GR00T + intent recognition). Without the intent module, the VLA must infer the interaction from its visual observations alone. A trial is counted as correct if the robot performs the interaction the partner initiated (Table~\ref{tab:online_f1}). All participants were informed about the experimental procedure and the laboratory’s internal regulations, which they acknowledged in writing prior to participation. Participation was voluntary, and participants were informed that they could withdraw from the experiment at any time.

GR00T is the best stand-alone VLA (macro-F1 0.75), but it often confuses the fist bump. Adding intent recognition raises macro-F1 to 0.90, improves every interaction (first bump: 0.50~$\rightarrow$~0.78), and eliminates errors on hugs (RQ3). The online ranking differs from the offline one: Being-H0.5 is competitive offline but almost never performs the requested interaction, while UnifoLM ranks second online. Open-loop error is thus a poor proxy for interaction performance, and on-robot testing is needed to select a VLA (RQ1). Both stop phrases were detected in 10/10 trials through the robot's onboard microphone at close range, with no false triggers during conversation (RQ3).

\begin{table}[t]
\centering
\caption{Online evaluation on the robot: F1 score per interaction (higher is
better), 150 trials per policy. \emph{Mean}: macro-F1 over the three
interactions. Best in bold.}
\label{tab:online_f1}
\small
\begin{tabular*}{\columnwidth}{@{\extracolsep{\fill}}lcccc@{}}
\toprule
 & \multicolumn{3}{c}{Interaction} & \\
\cmidrule(lr){2-4}
Model & Handshake & Fist bump & Hug & Mean \\
\midrule
Being-H0.5    & 0.09 & 0.00 & 0.26 & 0.12 \\
UnifoLM-VLA-0 & 0.57 & 0.38 & 0.94 & 0.63 \\
RLDX-1        & 0.68 & 0.38 & 0.54 & 0.53 \\
GR00T         & 0.86 & 0.50 & 0.89 & 0.75 \\
\midrule
GR00T + intent (ours) & \textbf{0.92} & \textbf{0.78} & \textbf{1.00} & \textbf{0.90} \\
\bottomrule
\end{tabular*}
\end{table}

\subsection{Interaction-Intent Model Comparison}
\label{sec:intent_model_comparison}

We evaluated six causal video classifiers formed by combining three lightweight visual encoders---TinyViT, MobileViT, and EfficientFormer --- with two temporal architectures, VTN and VideoLightFormer. All configurations used the same dual-view input, image resolution, and 16-frame history. The visual encoders were frozen during training, while the temporal modules, view-fusion layer, and classification head were optimized. The development data were derived from an approximately 1,000-episode dataset. This dataset was cleaned specifically for the interaction-phase prediction task by excluding episodes that were unsuitable for reliable phase prediction. After cleaning, 790 episodes remained, of which 632 were used for training and 158 for validation. A separate test dataset containing 288 episodes was used for the final model comparison. Because the test dataset contained no \texttt{no\_interaction} episodes, its macro-F1 was computed over the three available interaction classes: \texttt{handshake}, \texttt{hug}, and \texttt{fist\_bump}. Across the six configurations, test macro-F1 ranged from 0.51 to 0.76, with p95 inference latency ranging from 5.31 to 6.73\,ms. TinyViT with VideoLightFormer achieved the highest test macro-F1 of 0.76 and an accuracy of 0.736, while maintaining a p95 latency of 6.27\,ms. This configuration was therefore selected for integration with the VLA policy. On the validation set, the temporal commit rule produced no false activations across 54 \texttt{no\_interaction} episodes. Moreover, 77\% of interaction intents were committed before physical contact, with a median lead time of 2.1\,s. Full per-model results are provided in the supplementary material.

\subsection{Compliance User Study}
\label{sec:exp_compliance}

To answer RQ2, a within-subject study varies shoulder, elbow, and wrist stiffness between rigid and soft ($2\times2\times2$; soft: $K_p\times0.6$, $K_d\times0.83$), ordered by a balanced Latin square after an unrated rigid practice round. After each handshake, fist bump, and hug, participants rate safety, comfort, and naturalness on a 7-point scale. Two pilot sessions refined the protocol; data collection is ongoing.

\section{Results and Discussion}
\label{sec:results_discussion}

The offline results were similar for Being-H0.5, GR00T, and RLDX-1, with an overall RMSE between 0.133 and 0.135. However, their online performance was very different. GR00T achieved the highest stand-alone macro-F1 of 0.75, whereas Being-H0.5 reached only 0.12 despite its competitive offline error.
Thus, offline RMSE is not a reliable measure of real-robot interaction performance, and VLA models should also be evaluated online. For intent recognition, TinyViT with VideoLightFormer achieved the best results, with a test macro-F1 of 0.76 and a p95 latency of 6.27\,ms. It was therefore selected for integration with GR00T. The combined system improved GR00T's macro-F1 from 0.75 to 0.90. The largest gain was observed for fist bumps, where F1 increased from 0.50 to 0.78. The scores for handshakes and hugs also increased to 0.92 and 1.00, respectively. These results show that the intent module helps GR00T distinguish between similar interactions while maintaining real-time performance.

\subsection{Limitations and Future Work}

This study is limited by the number of participants, the use of only three predefined interaction types, limited environmental and user diversity, and evaluation on a single robot platform and whole-body controller. Although the intent dataset includes a \texttt{no-interaction} class, evaluation on labeled episodes does not fully represent continuous robot operation. Therefore, false activations during long periods without interaction require further study. The intent model may generalize poorly to unseen participants, especially for fist bumps. Speech recognition may also fail in noisy environments. Lowering $K_p$ gives no formal bound on contact forces and, without gravity compensation, lets the arms sag slightly; this was acceptable for these gestures but would limit precise low-stiffness tasks. Future work will expand the dataset, evaluate false activations during continuous operation, include more diverse users and environments, improve contact-aware perception, and investigate stronger physical safety guarantees.

\balance


\end{document}